\documentclass[runningheads]{llncs}
\usepackage{graphicx}
\usepackage{epstopdf}
\usepackage{amsfonts}
\usepackage{booktabs}
\usepackage{color}
\begin{document}
%
\title{REAPS: Towards Better Recognition of Fine-grained Images by Region Attending and Part Sequencing}
\titlerunning{REAPS: Towards Better Recognition of Fine-grained Images}
%
\author{
Peng Zhang\inst{1} \and
Xinyu Zhu\inst{2} \and
Zhanzhan Cheng\inst{1} \and
Shuigeng Zhou\inst{2} \and
Yi Niu\inst{1} 
}
\authorrunning{Peng et al.}
%
\institute{
Hikvision Research Institute, China 
\email{\{zhangpeng23,chengzhanzhan,niuyi\}@hikvision.com}\\
 \and
Fudan University, Shanghai, China\\
\email{\{16210720101,sgzhou\}@fudan.edu.cn}}

%
\maketitle              
\begin{abstract}
  Fine-grained image recognition has been a hot research topic in computer vision due to its various applications.
  The-state-of-the-art is the part/region-based approaches that first localize discriminative parts/regions, and then learn their fine-grained features.
  However, these approaches have some inherent drawbacks: 1) the discriminative feature representation of an object is prone to be disturbed by complicated background; 2) it is unreasonable and inflexible to fix the number of salient parts, because the intended parts may be unavailable under certain circumstances due to occlusion or incompleteness, and 3) the spatial correlation among different salient parts has not been thoroughly exploited (if not completely neglected).
  To overcome these drawbacks, in this paper we propose a new, simple yet robust method by building part sequence model on the attended object region.
  Concretely, we first try to alleviate the background effect by using a region attention mechanism to generate the attended region from the original image.
  Then, instead of localizing different salient parts and extracting their features separately, we learn the part representation implicitly by applying a mapping function on the serialized features of the object. 
  Finally, we combine the region attending network and the part sequence learning network into a unified framework that can be trained end-to-end with only image-level labels.
  Our extensive experiments on three fine-grained benchmarks 
  show that the proposed method achieves the state of the art performance.

  \keywords{Fine Grained  \and Region Attending \and Part Sequencing.}
\end{abstract}

\section{Introduction}
Fine-grained image recognition has attracted much research interest of the computer vision community \cite{angelova2013efficient,berg2013poof,berg2014birdsnap,zhang2012pose}, which tries to distinguish sub-ordinate categories such as car models~\cite{krause2015fine}, bird species~\cite{branson2014bird,huang2016part,WahCUB_200_2011}, dog breeds~\cite{khosla2011novel} and flower categories~\cite{nilsback2008automated} etc.
Though much effort \cite{fu2017look,huang2016part,jaderberg2015spatial,liu2016fully,wang2015multiple,zheng2017learning} has been devoted to solving this problem, recognizing fine-grained images is still a challenging task due to their relatively small inter-class difference and large intra-class variation.

\begin{figure}[th]
\centering
  \includegraphics[width=\textwidth]{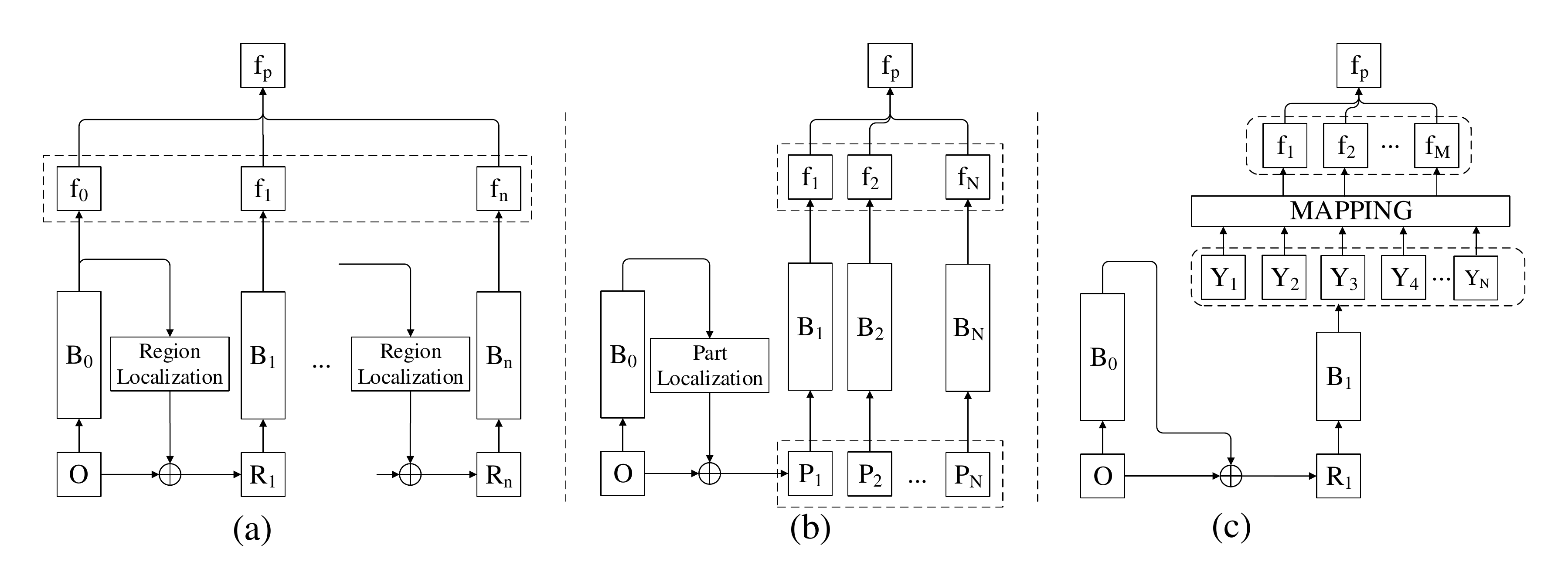}
\caption{An illustrative comparison between our framework and two popular existing fine-grained recognition frameworks.
        (a), (b) and (c) represent RLA, PL and our proposed framework respectively. $O$, $P_i$, $R_i$ and $B_i$ correspond to the original image, a selected part, an attended region and a backbone network respectively. \textcircled{+} represents the operation of crop and zoom in.}
\label{framework_vs}
\end{figure}

\begin{figure}[th]
\centering
  \includegraphics[width=\textwidth]{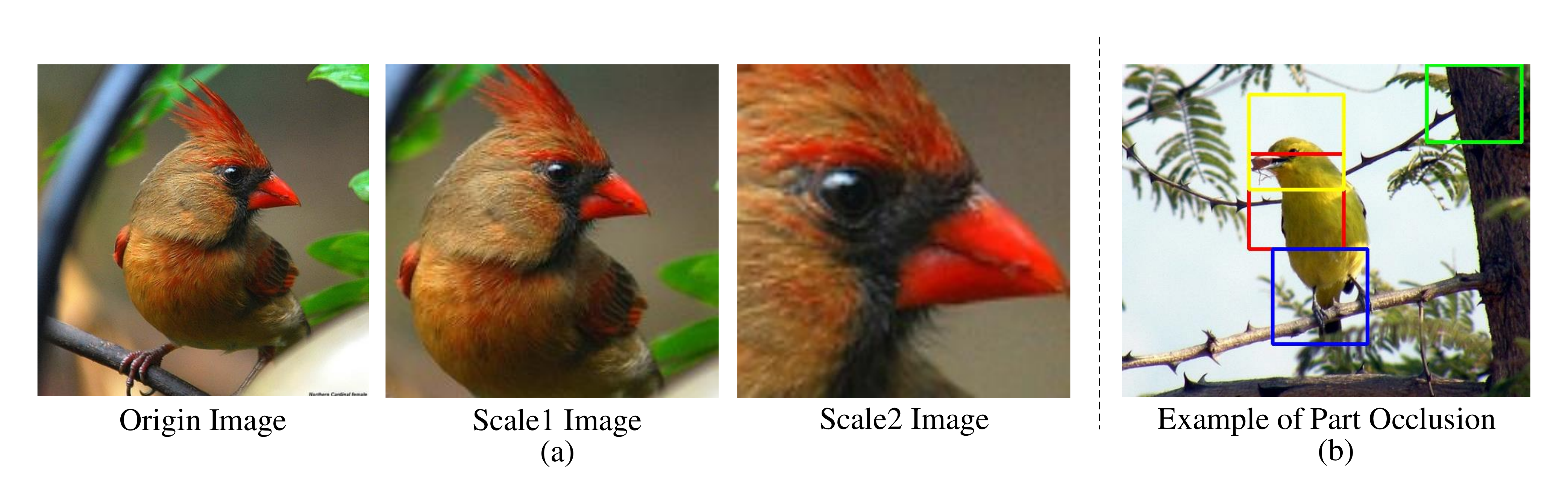}
\caption{Drawbacks of the RLA and PL frameworks.
    (a) RLA focuses on a detailed region progressively, while neglecting some other salient parts (the feet and wings of the bird disappear in scale1 and scale2 images).
    (b) PL detects a fixed number of pre-set parts and may get into trouble when some intended parts miss or be occluded (the back of the bird cannot be seen).
    [Best viewed in color]}
\label{limitations_pl_rl}
\end{figure}

Roughly speaking, there are two kinds of popular frameworks for handling fine-grained categorization: \emph{key region localization and amplification} (\emph{abbr}. RLA) and \emph{discriminative part learning} (\emph{abbr}. PL).
In general, RLA tries to attend and amplify the key region for capturing detailed visual representation while avoiding background disturbance. On the other hand,
PL usually first localizes discriminative parts via some sophisticated part selection mechanisms such as part attentions~\cite{fu2017look,wei2016mask,zheng2017learning} and convolutional responses~\cite{xiao2015application,zhang2016picking}, and then extracts the visual representations of the selected parts by using multiple independent feature extractors. Fig.~\ref{framework_vs} (a) and (b) illustrate these two frameworks. 
Though previous studies proved their effectiveness, they have several inherent drawbacks.
For example, RLA-based methods may miss some salient parts when progressively attending the key region, as shown in Fig.~\ref{limitations_pl_rl} (a).
PL-based methods usually fix the number of salient parts to be extracted, which is unreasonable and inflexible because in certain scenarios some of the intended parts may be unavailable due to image occlusion and incompleteness, as shown in Fig.~\ref{limitations_pl_rl} (b).
Furthermore, learning independent extractor for each salient part neglects the spatial correlation among these different parts, which should be useful for image recognition if properly exploited.
These drawbacks mentioned above motivate us to develop a new method for fine-grained image recognition by simultaneously taking region attending and part sequence learning into consideration. We call the new method \textbf{\emph{REAPS}} ------ an abbreviation of \textbf{RE}gion \textbf{A}ttending and \textbf{P}art \textbf{S}equence learning. The framework of REAPS is shown in Fig.~\ref{framework_vs} (c), and its detailed architecture is illustrated in Fig.~\ref{system}, which consists of two major components:

  \emph{\textbf{Region attending network}} (RAN): Inspired by previous works \cite{fu2017look,he2017fast}, we apply the class activation mapping (CAM) mechanism \cite{zhou2016learning} to constructing the region attending network for generating the region attention. The attended region cropped and amplified from the origin image are fed into the part sequence-learning network (PSN). RAN can effectively depress the impact of the background noise, while PSN can depict the detailed visual features well.

  \emph{\textbf{Part sequence-learning network}} (PSN):
        As mentioned above, the traditional part-based methods usually fix the number of parts and the representation of each part is explicitly learned by independent extractors. So we call them `hard-part'-based methods. In addition to the drawbacks we previously mentioned, `hard-part'-based methods adopt independent feature extractor (\emph{e.g.} VGG19) for each part, which incurs high computational cost.
%
        In order to overcome the above drawbacks, we propose the `soft-part' concept, which is implemented by mapping the serialized visual features into a group of implicit discriminative part representation and capturing the spatial correlation among different salient parts simultaneously. 

In REAPS, we integrate the region attending network and the part sequence learning network into a unified framework, which can be trained end-to-end with only image-level annotations.

Our contributions are as follows:
    1) We propose the novel `soft-part' concept, and implement this concept by designing a part sequence learning network (PSN), which learns implicit discriminative part representation and captures the spatial context simultaneously.
%
%
%
     2) We apply the region attending network to localizing the object region and alleviating the interference of complicated background to fine feature representation.
    3) We integrate the region attending network and the part sequence learning network into a unified framework, and train it end-to-end without any part-level annotation.
    4) We conduct extensive experiments on three challenging datasets (Stanford Cars, FGVC-Aircraft and CUB Birds), which demonstrate the superiority of our method over the existing ones.

\section{Related Work}\label{gen_inst}

Fine-grained recognition (or categorization) is a challenging problem that has been extensively studied. Related works can be grouped in two dimensions: representation learning and part localization.

\subsection{Representation Learning}
Discriminative representation learning is crucial for fine-grained recognition.
Thanks to their strong encoding capability, most existing fine-grained recognition algorithms \cite{cui2017kernel,fu2017look,moghimi2016boosted,zheng2017learning} employ deep convolutional networks for feature representation, which have achieved much better performance than traditional descriptors and hand-crafted features \cite{perronnin2010improving,zhang2013deformable}.

To better handle the subtle inter-class difference and large intra-class variation in fine-grained recognition tasks, \cite{lin2015bilinear} proposes a bilinear structure to model second-order interactions of local convolutional features in a translationally invariant manner.
This idea was later extended by \cite{lin2017improved} and other variants \cite{gao2016compact,kong2017low}for better recognition performance.
Recently, \cite{cai2017higher,cui2017kernel} further exploit higher-order integration of convolutional activations that can yield more discriminative representation, and achieve impressive performance.

Besides, some approaches (e.g. \cite{qian2015fine}, \cite{zhang2016embedding}) try to learn more robust representations via distance metric learning. \cite{zhang2016picking} unifies deep CNN features with spatially weighted Fisher vectors to capture important details and eliminate background disturbance. \cite{moghimi2016boosted} incorporates deep CNNs into a generic boosting framework to combine the strength of multiple weaker learners, which improves the classification accuracy of a single model and simplifies the network design.

\subsection{Part Localization}
Previous works have studied the localization impact of discriminative parts on capturing subtle visual difference.
Early part-based approaches rely on extra annotations of bounding boxes or part landmarks to localize pre-defined semantic parts.
\cite{gavves2013fine,liu2012dog} assume that annotations are available in both training and testing.
Some later works \cite{huang2016part,krause2015fine,lin2015deep} use annotations only in training.
However, the cost-prohibitive manually-labeled annotations prevent the application of these algorithms to large-scale real problems.
Therefore, most of recent works focus on weakly-supervised task-driven part localization with only category labels.
Attention-based models have been widely used to automatically discover salient parts.
\cite{xiao2015application} proposes a two-level attention model, where one object-level filter-net selects relevant patches for a certain object, another part-level domain-net localizes discriminative parts.
Since deep filter responses from CNNs are able to significantly and consistently respond to specific visual patterns, \cite{zhang2016picking} proposes to learn part detectors by picking distinctive filters, while \cite{wang2015multiple} identifies discriminative regions according to channel activations.
\cite{jaderberg2015spatial} takes one step further and proposes a spatial transform module based on the differentiable attention mechanism, which enables CNN to learn better invariance to classification and all kinds of warping.

The latest works use hidden filter responses of deep CNNs as part detectors.
\cite{liu2016fully} proposes a fully convolutional attention network to optimally glimpse local discriminative regions by reinforcement learning.
\cite{fu2017look} introduces a recurrent attention convolutional neural network (RA-CNN) that recursively learns discriminative region attention from coarse to fine by an attention proposal network.
\cite{zheng2017learning} develops a multi-attention convolutional neural network (MA-CNN) that generates multiple part attentions by clustering, weighting and pooling from spatially-correlated channels, and achieves the state-of-art performance.



\section{The REAPS Method}
\subsection{Overview}
\begin{figure}[t]
  \centering
  \includegraphics[width=\textwidth]{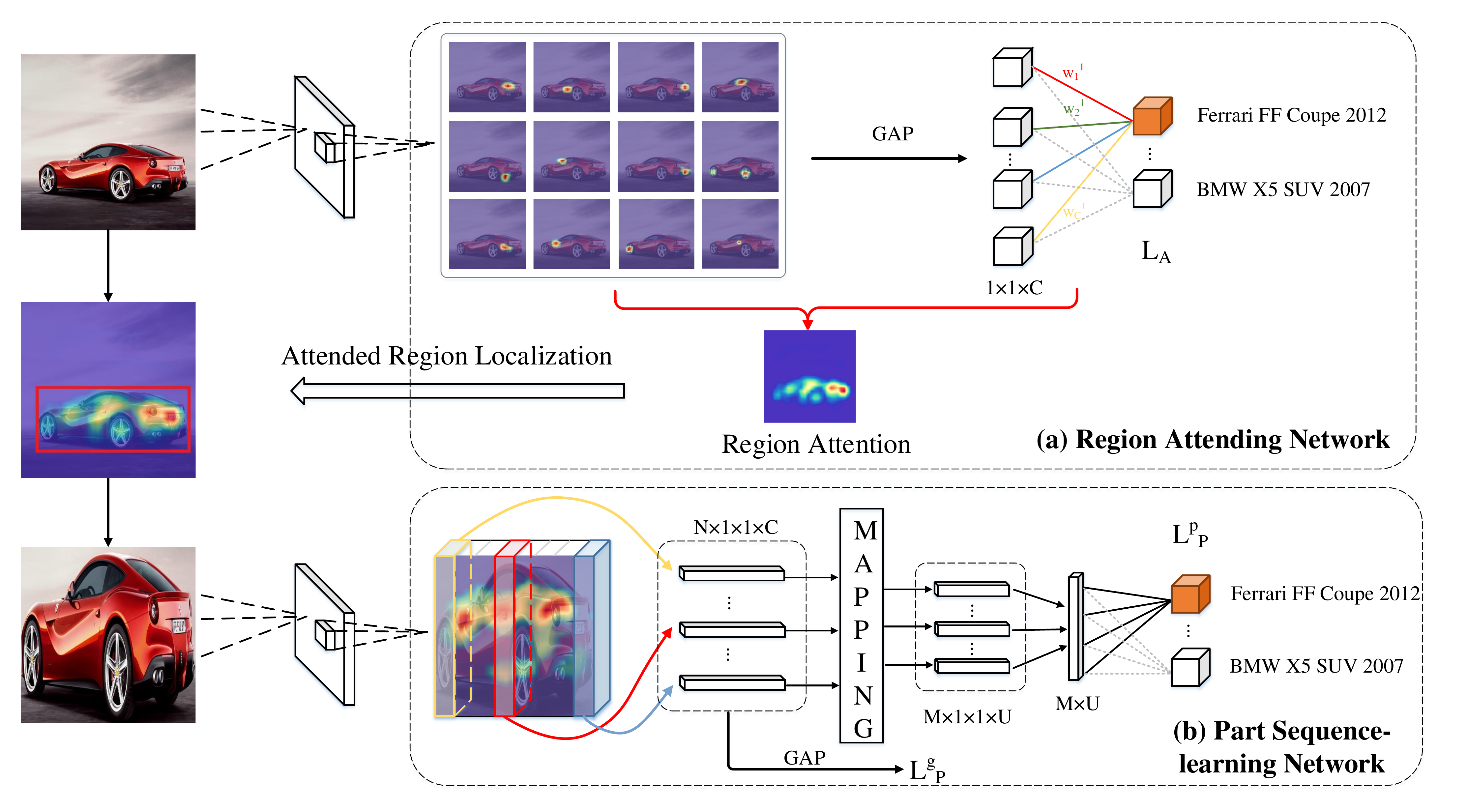}
  \caption{The REAPS architecture. The region attending network (RAN) takes an original image as input and produces the region attention by weighting the last convolutional feature maps with the parameters of softmax layer. The attended region is cropped out and zoomed in, and fed into the part sequence learning network (PSN) where part representation is learned in an implicit way by applying the mapping function on the serialized features. The whole network can be trained end-to-end under the supervision of three softmax loss functions in Eq.~(\ref{loss_function}). [Best viewed in color]}
  \label{system}
\end{figure}

The architecture of our REAPS method is shown in Fig.~\ref{system}, which consists of two major components: a \emph{region attending network}~(RAN) and a \emph{part sequence learning network}~(PSN).
RAN leverages deep convolutional responses to generate the discriminative region attention.
Then the attended region is cropped out and zoomed in as the input of PSN.
While in PSN, the deep visual features are extracted by a backbone network and further serialized to a sequence of vectors, each of which describes a rectangle region in the raw image. A mapping function is learned to map the sequence of vectors to discriminative part representation.

\subsection{Region Attending Network}

The RAN is based on a standard classification network, where global average pooling is applied on the last convolutional feature maps, followed by a SoftmaxLoss layer. Given an input image $\mathcal{I}$, we can get its last convolutional feature maps, denoted as $\mathcal{F}$. 
In order to eliminate the background disturbance, we apply the CAM \cite{zhou2016learning} mechanism to generating the region attention in a ``self-guided'' way.
Concretely, let $f_{(x,y)}^k$ denote the activation value of unit $k$ at $(x, y)$ in $\mathcal{F}$, the global average pooling (\emph{abbr.} GAP) can be represented as $\mathcal{P}_k=\sum_{x,y} f^k_{(x, y)}$.
For a specific class $c$, the probability yielding $c$ is $\sum_k W_{k, c} \mathcal{P}_k$, where $W_{k, c}$ is the weight of the last inner-product layer and can be learned under the supervision of SoftmaxLoss $\mathcal{L}_A$.
$W_{k, c}$ also acts as the weight indicating the importance of unit $k$ for class $c$.
Therefore, we can compute the region attention of $c$ at $(x, y)$ by
\begin{equation}
RA^c_{x,y} = \sum_k W_k^c f^k_{(x, y)}.
\end{equation}

Since each unit $k$ responds to a certain type visual pattern (\emph{e.g.} circle), the region attention can represent all the discriminative visual patterns at different locations by conducting weighted-sum over all units.
Furthermore, we locate the attended region of class $c$ by

\begin{equation}
\label{cam_op}
AR^c = f_{+}(BB_\tau(RA^c), \mathcal{I}),
\end{equation}

where $BB_\tau$ is the operation of calculating rectangular bounding box over the binary mask based on the preset threshold $\tau$, and $f_{+}$ represents the operation of crop and zoom in.


\subsection{Part Sequence Modeling}
\label{sec_part_model}

Let $X\in \mathbb{R}^{(H\times W\times C)} $ denote the deep representation through the deep convolutional neural network (the backbone network in Fig.~\ref{system}), where $H$, $W$ and $C$ respectively refer to the height, width and the number of channels of $X$.
Instead of localizing a fixed number of parts and extracting their representation separately, we learn the part representation in a soft way. Concretely, we evenly decompose $X$ into a sequence of $N$ vectors by
\begin{equation}
Y = [Y_1, Y_2, ..., Y_N] = seq(X),
\end{equation}
where $Y_i \in \mathbb{R}^{(1\times C)}$ describes a rectangle region in the raw image and $seq$ is the pooling operation with kernel size of $[H\times\frac{W}{N}]$.
%
%
We abstract the sequence of visual feature vectors into $M$ implicit parts with the learned mapping function
\begin{equation}
P_P = [P_1, P_2, ..., P_M] \simeq mapping(Y),
\end{equation}
%
where $M \leq N$ refers to the rough number of learnt discriminative parts.

The mapping function here should keep the discriminative features and depress the useless ones. Several sequence learning techniques meet this requirement, \emph{e.g.} the recurrent neural networks (GRU~\cite{cho2014properties} and LSTM~\cite{hochreiter1997long}) and attention-based sequence models~\cite{cho2014learning,xu2015show}.
Attention-based models can fulfill the mapping from a source sequence of length $N$ to a target sequence of length $M$. With sequential labels, they can effectively learn the alignment between labels and their corresponding salient representation of features~\cite{cho2014learning,xu2015show}.
Recurrent neural networks (GRU~\cite{cho2014properties} and LSTM~\cite{hochreiter1997long} can strengthen the sequence representation with the sequence length being unchanged. 
%
In our case, classification is performed in a weakly-supervised fashion and the only available supervision information is the image-level category, 
so we apply a bi-directional LSTM on the sequenced vectors and concatenate all the hidden states as the part representation $P_P\in \mathbb{R}^{(N\times U)} $, followed by a fully connected layer and a SoftmaxLoss $\mathcal{L}_P^p$.

To accelerate the training process, another SoftmaxLoss $\mathcal{L}_P^g$ is attached to the global representation $P_g\in \mathbb{R}^C$ of the backbone network of PSN.

\subsection{Training and Joint Representation}

Instead of alternative optimization, REAPS can be trained end-to-end straightforwardly by
\begin{equation}
\mathcal{L}=\lambda_1\mathcal{L}_A + \lambda_2\mathcal{L}_P^g + \lambda_3\mathcal{L}_P^p,
\label{loss_function}
\end{equation}
where $\lambda_j(j = 1,2,3)$ is the corresponding loss weight.
Once the training process converges, the joint representation $F$ of the input image $\mathcal{I}$ can be represented by a set of descriptors, followed by a fully-connected layer with softmax function for final classification:
\begin{equation}
F=\{A_g, P_g, P_p\},
\end{equation}
where $A_g\in \mathbb{R}^C$ denotes the global representations of the backbone network of RAN.



\section{Performance Evaluation}

\subsection{Datasets and Implementation Details}

We conduct extensive experiments on three benchmark datasets, including Stanford Cars\cite{krause2015fine}, FGVC-aircraft\cite{maji2013fine} and CUB-200-2011\cite{WahCUB_200_2011}, which are widely used to evaluate fine-grained image recognition. Table~\ref{dataset} shows the detailed statistics of the three datasets.

\begin{table}
  \caption{Statistics of the fine-grained benchmark datasets used in this paper.}
  \label{dataset}
  \centering
  \begin{tabular}{llll}
    \toprule
    Dataset          &  Category     &  Training     &  Testing\\
    \midrule
    Stanford Cars\cite{krause2015fine} & 196           & 8,144         & 8,041   \\
    FGVC-aircraft\cite{maji2013fine} & 100           & 6,667         & 3,333   \\
    CUB-200-2011\cite{WahCUB_200_2011}  & 200           & 5,994         & 5,794   \\
    \bottomrule
  \end{tabular}
\end{table}

For fair comparison, all compared methods employ similar backbone network. Specifically, we start with the 19-layer VGGNets pre-trained on ImageNet and fine-tune it on the three fine-grained datasets. The parameters of RAN and PSN are initialized with the same pre-trained model. Input images and the cropped attended regions are both resized to $448\times448$, where high resolution highlights details and benefits recognition. We use SGD with momentum 0.9 to minimize the loss function $\mathcal{L}$ in Eq.~(\ref{loss_function}), where $\lambda_1$, $\lambda_2$, and $\lambda_3$ are all set to 1. The threshold $\tau$ in Eq.~(\ref{cam_op}) is set to $0.1$. Following the common practice of learning rate decaying schedule, the initial learning rate is set to 0.001 and multiplied by 0.1 every 60 epoches.

\subsection{Experiment and Analysis}

\begin{figure}
  \centering
  \includegraphics[width=\textwidth]{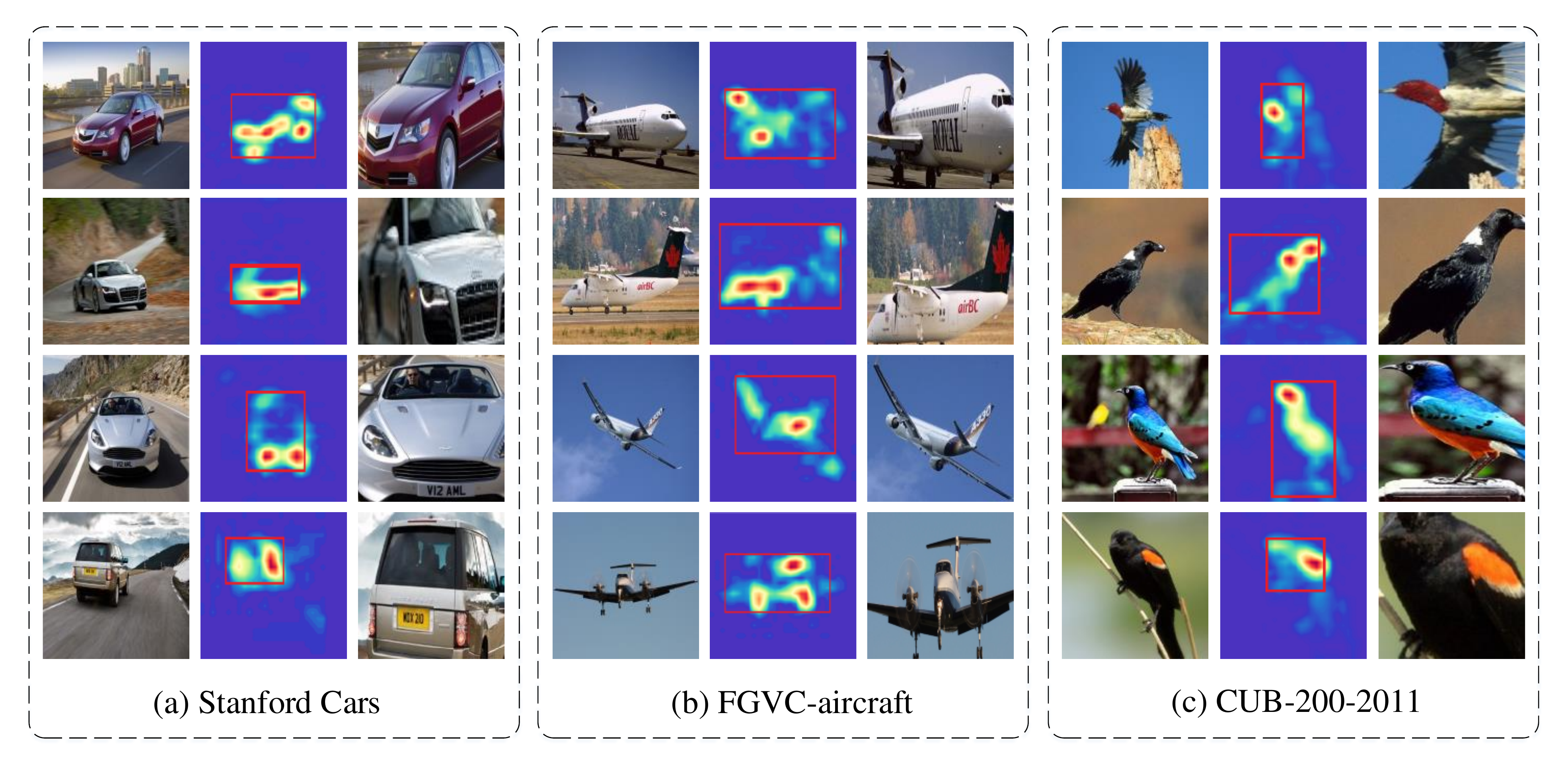}
  \caption{Region attention localization results of RAN for some examples from (a) Stanford Cars, (b) FGVC-aircraft, and (c) CUB-200-2011. Pictures from left to right in (a-c) are the raw image, the attention mask with bounding box indicating the area of top attention response, and the cropped object-level region respectively.}
  \label{region-atten-fig}
\end{figure}

\paragraph{Effectiveness of Region Attending Network.} We first visualize some results of our CAM-based region attending network in Fig.~\ref{region-atten-fig} for qualitative analysis. We can see that the discriminative regions of the input images are highlighted. The attended regions that are cropped from the raw images and then amplified preserve the object-level structure, eliminate background interference and enrich local visual details.
We evaluate the effectiveness of RAN in terms of the single scale classification accuracy. In Table~\ref{region-atten-table}, we compare the result of our PSN \emph{without} part modeling branch with that of two other attention based approaches. To be fair, we select the \emph{single-attention} based performance from FCAN~\cite{liu2016fully}, and \emph{the second scale} result from RA-CNN~\cite{fu2017look}. RA-CNN\cite{fu2017look} is the most relevant work to ours considering the region attention concept and the way to use it. We can see that our method outperforms FCAN~\cite{liu2016fully} with a clear margin (7.1\% relative gain) and RA-CNN\cite{fu2017look} with 1.3\% accuracy improvement

\begin{figure}
  \centering
  \includegraphics[width=\textwidth]{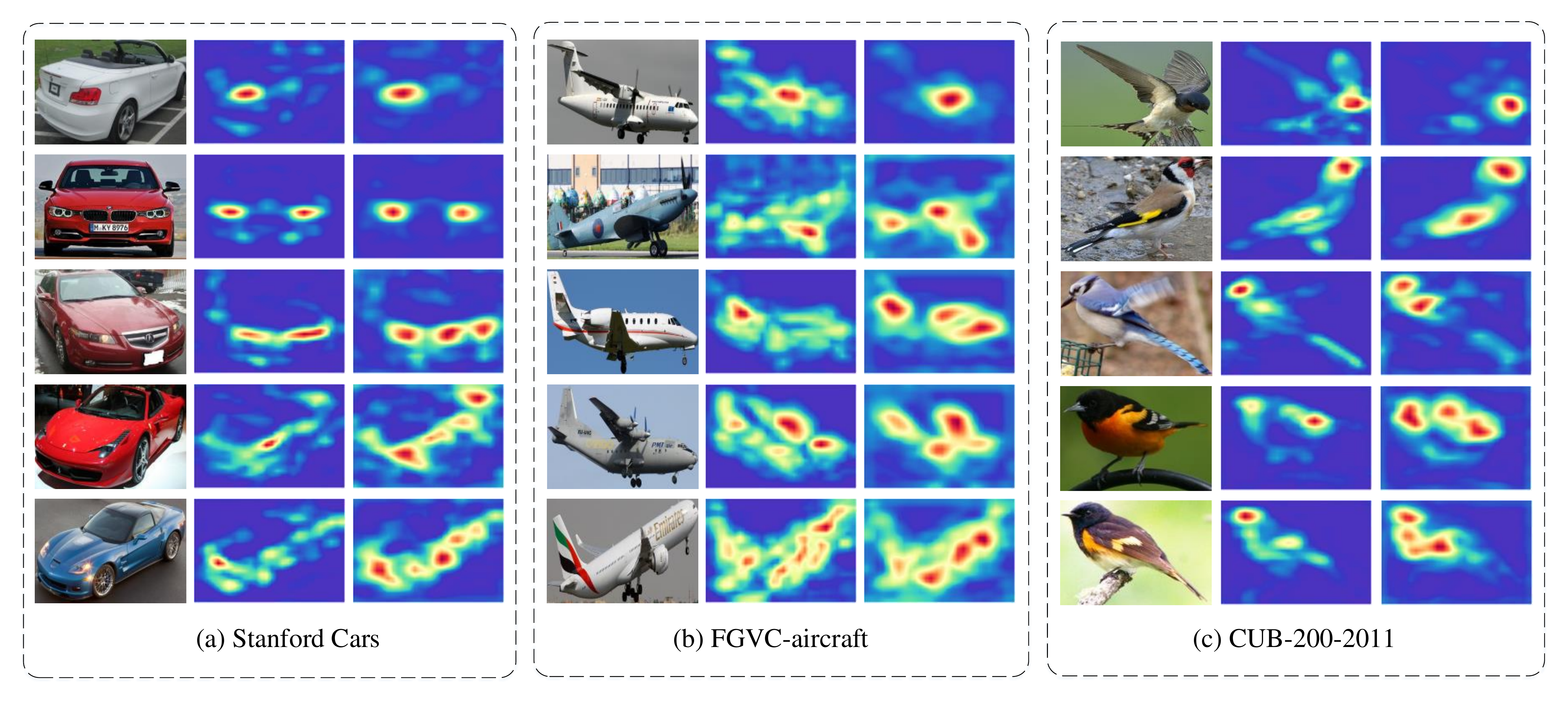}
  \caption{Visualization of feature maps for some examples from (a) Stanford Cars, (b) FGVC Aircraft, (c) CUB-200-2011. Pictures from left to right in (a-c) are the raw image, the feature map generated by PSN \emph{without} part branch and the feature map generated by PSN \emph{with} part branch, respectively.}
  \label{lstm-part-fig}
\end{figure}

\paragraph{Effectiveness of part branch in PSN.}
As pointed out in Section~\ref{sec_part_model}, we take a bi-directional LSTM as the mapping function to map the serialized features to discriminative part representation.
To evaluate its effectiveness, we present the classification results of two models: \emph{PSN wo/w part}, which are the classification accuracy based on the $P_g$ \emph{without/ with} the part modeling branch. The results are shown in Table~\ref{region-atten-table}. We can see that the model adopting part sequence modeling branch achieves a relative performance gain of 1.0\%. Some illustrations are given in Figure~\ref{lstm-part-fig}.
It can be observed that the part branch encourages a more compact distribution on the feature maps, further enhances the crucial part areas and depresses the unless ones.

\begin{table}
  \caption{Performance comparison of attention localization on the Stanford Cars dataset.}
  \label{region-atten-table}
  \centering
  \begin{tabular}{cc}
    \toprule
    Approach                                        &  Accuracy\\
    \midrule
    FCAN (single-attention)~\cite{liu2016fully}      &  84.2    \\
    RA-CNN (scale 2)~\cite{fu2017look}               &  90.0    \\
    PSN \emph{wo part}                        &  91.3    \\
    PSN                                   &  92.3    \\
    \bottomrule
  \end{tabular}
\end{table}

\paragraph{Fine-grained Categorization.}
We compare our REAPS framework with several existing methods and the results are summarized in Table~\ref{exp-summary}. Our method achieves better performance than those~\cite{branson2014bird,krause2015fine,wang2015multiple,wang2016mining,wei2016mask} using ground-truth bounding boxes or part annotations during training or testing time on three datasets. Compared with BCNN-based methods~\cite{cui2017kernel,lin2017improved,moghimi2016boosted} 
our method obtains comparable or even better performance due to accurate attention localization and part sequence modeling.
To further enhance the capacity of REAPS, as RA-CNN~\cite{fu2017look} does, we incorporate one more PSN into our framework, and the 2nd PSN is based on the attended region of the 1st PSN. We call the resulting network REAPS+.
Note that, REAPS+ obtains the best performance on three datasets. Especially on FGVC Aircraft dataset, our proposed REAPS+ obtains the best accuracy of 92.6\%, surpassing state-of-the-art MA-CNN~\cite{zheng2017learning} by a relative 2.7\% gain. The significant improvement suggests that the proposed part sequence modeling network works as expected to leverage spatial information of parts, and does even better when the objects to be recognized have strong sequential structures.



\begin{table}
\centering
\caption{Performance comparison on the Stanford Cars, FGVC Aircraft and CUB-200-2011 datasets. \emph{(*)} indicates whether bounding box or part annotation is used in training.}
\label{exp-summary}
\begin{tabular}{l|c|c|c}
\hline
Approach            & Stanford Cars & FGVC Aircraft & CUB200-2011 \\ \hline
PA-CNN~\cite{krause2015fine}              & 92.8 (*)      & $-$           & 82.8 (*)    \\
MDTP~\cite{wang2016mining}                & 92.5 (*)      & 88.4 (*)      & $-$         \\
MG-CNN~\cite{wang2015multiple}              & $-$           & 86.6 (*)      & 83.0 (*)    \\
PN-CNN~\cite{branson2014bird}              & $-$           & $-$           & 85.4 (*)    \\
Mask-CNN~\cite{wei2016mask}            & $-$           & $-$           & 85.4 (*)    \\
STNs~\cite{jaderberg2015spatial}                & $-$           & $-$           & 84.1        \\
FCAN~\cite{liu2016fully}                & 91.5          & $-$           & 84.3        \\
PDFR~\cite{zhang2016picking}                & $-$           & $-$           & 84.5        \\
Improved B-CNN~\cite{lin2017improved}      & 92.0          & 88.5          & 85.8        \\
BoostCNN~\cite{moghimi2016boosted}            & 92.1          & 88.5          & 86.2        \\
KP~\cite{cui2017kernel}                  & 92.4          & 86.9          & 86.2        \\
RA-CNN(scale 1+2+3)~\cite{fu2017look} & 92.5          & $-$          & 85.3        \\
MA-CNN~\cite{zheng2017learning}              & 92.8          & 89.9          & \textbf{86.5}        \\ \hline
REAPS \emph{wo PSN}        & 92.0          & 89.8        & 81.3     \\
REAPS               & \textbf{93.1}          & \textbf{91.8}          &  86.0       \\
REAPS+              & \textbf{93.5}          & \textbf{92.6}           &  \textbf{86.8}       \\ \hline
\end{tabular}
\end{table}

\section{Conclusion}
In this paper, we propose a novel framework REAPS for fine-grained recognition, which consists of a region attending network and a part sequence-learning network. The proposed framework does not need bounding box/ part annotations for training and can be trained end-to-end. We conduct extensive experiments on three fine-grained benchmark datasets, and the experimental results show that REAPS outperforms the existing methods.

%
%
%
%
\bibliographystyle{splncs04}
\bibliography{mycitation}
\end{document}